\documentclass[10pt]{article}
\usepackage[preprint]{tmlr}   

\usepackage{graphicx}
\usepackage{booktabs}
\usepackage{multirow}
\usepackage{xcolor}
\usepackage{amsmath,amssymb}
\usepackage{enumitem}
\usepackage{tcolorbox}
\tcbuselibrary{breakable}

\usepackage{amsmath,amsfonts,bm}

\def\eqref#1{equation~\ref{#1}}

\def\1{\bm{1}}

\DeclareMathAlphabet{\mathsfit}{\encodingdefault}{\sfdefault}{m}{sl}
\SetMathAlphabet{\mathsfit}{bold}{\encodingdefault}{\sfdefault}{bx}{n}

\usepackage{hyperref}
\usepackage{url}

\newcommand{\bench}{MissingBench-Verified}
\newcommand{\nv}{\textsc{nv}}
\newcommand{\hs}{\textsc{hs}}
\newcommand{\vis}{\textsc{v}}
\newtcolorbox{promptbox}[1]{colback=gray!4,colframe=gray!60,fonttitle=\bfseries\small,title=#1,fontupper=\ttfamily\footnotesize,left=4pt,right=4pt,top=3pt,bottom=3pt,breakable}
\newtcolorbox{figbox}[1]{colback=gray!4,colframe=gray!60,fonttitle=\bfseries\scriptsize,title=#1,fontupper=\ttfamily\scriptsize\raggedright,left=3pt,right=3pt,top=2pt,bottom=2pt,boxsep=1pt,before skip=3pt,after skip=3pt}

\title{MissingBench-Verified: Vision-Language Models Often Report a Removed Essential Part as Visible}

\author{\name Guoping Luo$^{*}$ \email gluo06@student.ubc.ca \\
      \addr Weathon Software\\
      Department of Computer Science\\
      University of British Columbia
      \AND
      \name Wenqi Marshall Guo$^{*}$ \email wg25r@student.ubc.ca \\
      \addr Department of CMPS\\
      University of British Columbia - Okanagan Campus\\
      Weathon Software
      \AND
      \name Qingyun Qian \email qingyunq@student.ubc.ca \\
      \addr Department of CMPS\\
      University of British Columbia - Okanagan Campus\\
      Weathon Software
      \AND
      \name Shiyu Zhou \email szhou49@student.ubc.ca \\
      \addr Weathon Software\\
      Department of Computer Science\\
      University of British Columbia
      \AND
      \name Mohamed S. Shehata \email mohamed.sami.shehata@ubc.ca \\
      \addr Department of CMPS\\
      University of British Columbia - Okanagan Campus
      \AND
      \name Shan Du \email shan.du@ubc.ca \\
      \addr Department of CMPS\\
      University of British Columbia - Okanagan Campus
      \AND
      \addr $^{*}$Equal contribution.}

\def\month{MM}
\def\year{YYYY}
\def\openreview{\url{https://openreview.net/forum?id=XXXX}}

\begin{document}

\maketitle

\begin{abstract}
Inspection often asks whether a part is \emph{absent}, whereas vision-language models (VLMs) are mostly evaluated on what is \emph{present}. We introduce \bench{}, a benchmark of 452 real photographs in which exactly one functionally essential part of an object has been removed, each paired with its unedited original and verified by three annotators who answer the same question as the models and are not told which images were edited (98\% of their individual answers on edited images are correct). Nine widely deployed proprietary and open-weight VLMs identify the removed part as ``not visible'' on 40--70\% of edited images while answering the originals correctly 94--100\% of the time; five of the nine are below 50\%, and 73 items are missed by all nine. A simulated detector report stating that the part is undetected raises this rate on every model, by more when it arrives after the model has answered than before the question (9 of 9), yet 44--85\% of the initial errors remain. An open-ended framing that never names the part gives the same ranking of the five models tested, and independent runs move no model's rate by more than 4 points. Three open-vocabulary detectors queried with the part name separate edited from original images poorly (AUROC $\leq 0.64$). In one open-weight model, Qwen3-VL-8B, a linear probe on the hidden state separates edited from original images at an AUROC of 0.89--0.94 from the second layer on, better than the model's own answer (0.82), although the probe cannot tell the missing part from traces of the edit.
\end{abstract}

\section{Introduction}
\label{sec:intro}

Maintenance checklists, quality-control stations and safety audits often ask whether something is absent. Is the pressure-gauge needle there? Does the wheelchair still have its front wheels? Has the guard been removed from the saw? Vision-language models are increasingly placed in these roles, while most benchmarks measure their competence on what an image \emph{contains}. \bench{} measures the complementary ability: given a real photograph of an ordinary object and the name of one of its essential parts, can the model tell that the part is not there?

\begin{figure}[t]
    \centering
    \includegraphics[width=\linewidth]{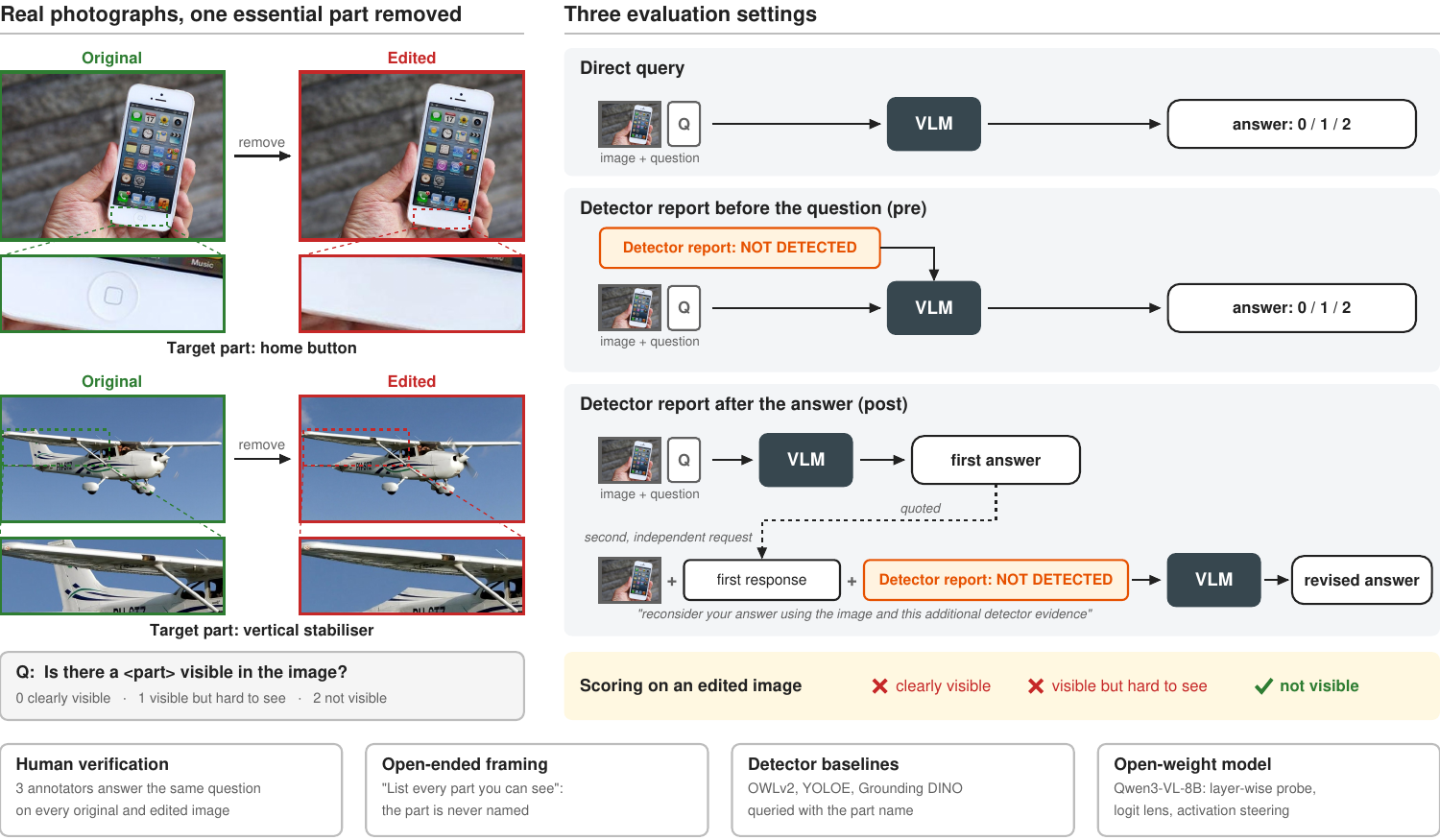}
    \caption{Overview of \bench{}. Left: two items, each a real photograph and its edited copy from which one essential part has been removed (dashed boxes mark the enlarged region). Right: the three evaluation settings of Section~\ref{sec:protocol}, in which the model answers the three-way question directly, with a simulated detector report placed before the question, or with the same report delivered in a second request after its first answer; on an edited image only ``not visible'' is correct. Bottom: the further analyses of Sections~\ref{sec:benchmark}, \ref{sec:results} and~\ref{sec:mechanism}.}
    \label{fig:teaser}
\end{figure}

On 452 verified items, nine widely deployed VLMs released in 2026 answer ``not visible'' between 40\% and 70\% of the time, five of them below 50\% (Table~\ref{tab:main}). The same models answer the unedited originals correctly between 94\% and 100\% of the time. Three annotators, asked the same question without being told which images were edited, answer ``not visible'' on 98\% of edited images. Seventy-three items are answered incorrectly by every model.

Every item is a real photograph, selected by hand, in which exactly one \emph{functionally essential} part has been removed with a generative editor or a manual eraser; the object cannot perform its function without the part (an aircraft without a vertical stabiliser, a bicycle without a handlebar, a kettle without a spout). Every item is paired with its original, which serves as the per-item control for the question wording, the image quality and the model's knowledge of the part. Several items share a source photograph with different parts removed. The resulting set covers 387 distinct part names across household objects, vehicles, tools, instruments and electronics.

The benchmark also quantifies how much an explicit report changes an answer. A simulated detector report that the part is undetected raises the ``not visible'' rate on every model, and raises it more when delivered after the model has answered than when delivered before the question (post exceeds pre on 9 of 9 models, sign test $p=0.004$). Even then, 44--85\% of a model's initial errors remain (Section~\ref{sec:evidence}). The measurement is stable: an open-ended framing that never names the part gives the same ranking of the five models tested, and two independent runs of the direct query move no model's rate by more than 4 points (Section~\ref{sec:results}).

We then ask where the error arises, using three open-vocabulary detectors and one open-weight VLM (Section~\ref{sec:mechanism}). The detectors, queried with the part name, separate edited from original images at an AUROC of 0.64 or below. In Qwen3-VL-8B, run with the same prompt, a linear probe on the language model's hidden state separates edited from original images at 0.89--0.94 from the second layer to the last, whereas the model's own answer separates them less well: its answer without reasoning (the \emph{immediate answer}) reaches 0.82, and its answer after reasoning is correct on 29\% of edited images. Steering the hidden state along the class-mean direction raises the balanced accuracy of the reasoned answer from 62.6\% to 71.7\%. These results are consistent with an error that arises, in this model, in the \emph{readout}, the computation that turns the hidden state into an answer. They do not establish it, because the probe is trained on edited versus original images and may read traces of the edit instead of the absence of the part.

\paragraph{Contributions.}
\begin{itemize}[leftmargin=1.2em, itemsep=1pt]
    \item \textbf{\bench{}}: 452 human-verified real-photo items with one essential part removed, each paired with its original, released with metadata, verbatim prompts, every per-item model output and the evaluation code (Section~\ref{sec:benchmark}).
    \item \textbf{A measurement}: nine deployed VLMs detect the absence on 40--70\% of items with originals at 94--100\%, stable across two framings and two runs, with a protocol that scores how much a report delivered before or after the answer changes it (Section~\ref{sec:results}).
    \item \textbf{An analysis of one open-weight model}: depth probes, a logit lens, activation steering and detector baselines, which show that the hidden state separates edited from original images better than the answer does (Section~\ref{sec:mechanism}).
\end{itemize}

\section{Related Work}
\label{sec:related}

\paragraph{Object hallucination benchmarks.}
Object hallucination, an answer that names an object the image does not contain, has been measured since CHAIR \citep{rohrbach_object_2019} and, for instruction-tuned models \citep{liu_visual_2023,liu_improved_2024}, by polling ``Is there a bottle in the image?'' \citep{li_pope_2023}. Later suites broaden the questions and metrics \citep{wang_amber_2023,guan_hallusionbench_2024,liu_survey_2024}, pose questions whose correct answer is ``no'' \citep{lovenia_nope_2023}, or insert and delete objects synthetically \citep{wu_autohallusion_2024}. Two lines come closest to an edited-image design: BEAF pairs each image with a version from which an object has been removed and scores the change in the answer \citep{ye-bin_beaf_2025}, and SegSub substitutes or removes segmented regions to test whether a model follows the pixels or its expectation \citep{carragher_segsub_2025}. The objects in these benchmarks are whole and separable, and their removal leaves the scene plausible. \bench{} asks the absence question about one functionally essential part of a single photographed object, with the unedited original as the per-item control.

\paragraph{Prior-driven bias.}
A second line of work shows that answers follow what an object usually has rather than what the pixels show. Familiar objects altered by one element, an animal with an added leg or a logo with an extra stripe, are counted as the canonical object \citep{vo_vlmbias_2025}; MM-SpuBench measures how far answers track spurious correlations between an object and its context \citep{ye_mm-spubench_2025}; WHOOPS! collects synthetic scenes that violate commonsense and finds that models explain the violation away \citep{bitton_whoops_2023}. These studies mostly use rendered or synthetic images, or counts and textures of a whole object; our items are edited real photographs in which the altered element is a part implied by the object's function.

\paragraph{Knowledge conflict and sycophancy.}
When retrieved text contradicts a language model's parametric knowledge, the model's choice between the two depends on its confidence in each \citep{wu_clasheval_2025}, and the framing of the context decides which side wins \citep{jin_tug--war_2024}. Sycophancy studies measure a related pressure: a model revises a correct answer when a user pushes back, and preference models reward the revision \citep{sharma_sycophancy_2024}. These results concern text evidence against text knowledge, or a human challenger whose claim is unverified. In our protocol the evidence is a fixed tool report whose content is true for the image, delivered either before the question or after the model has answered.

\paragraph{Missing-component inspection.}
Industrial inspection treats a missing part as one anomaly class among scratches, dents and contamination: MVTec AD includes missing-component defects for some product categories, photographed under fixed studio conditions \citep{bergmann_mvtec_2019}, and MMAD evaluates multimodal models on question answering over such images \citep{jiang_mmad_2024}. A part can also be sought with an open-vocabulary detector: a text query for the part name returns boxes and confidences from OWLv2 \citep{minderer_scaling_2024}, Grounding DINO \citep{liu_groundingdino_2024} or YOLOE \citep{wang_yoloe_2025}, and a low confidence is read as absence. We run these three detectors on the same original--edited pairs as a baseline (Section~\ref{sec:detectors}).

\paragraph{Representation probing and activation steering.}
Linear probes read a quantity from intermediate activations and locate the depth at which it is linearly accessible \citep{alain_probes_2016}; the logit lens applies the unembedding to every layer without training \citep{nostalgebraist_logitlens_2020}; activation steering adds a direction to the residual stream at inference. These tools were developed on text-only models and have since been applied to VLMs, where the closest results to ours concern evidence that is inside the model but does not reach the answer. \citet{seeing_not_believing} report that VLMs often attend to the correct visual evidence in their deeper layers even when they answer wrongly, and \citet{encoded_but_unused} separate errors in which the visual evidence is not encoded from errors in which it is encoded but not used, estimating that the second kind accounts for most errors of two open-weight VLMs. Our analysis asks the same question about an absent part in one open-weight model, with the originals as the paired control: a depth probe at the last prompt token, a logit lens, the model's answer with and without chain of thought \citep{wei_chain--thought_2023}, and steering.

\section{\bench{}}
\label{sec:benchmark}

\subsection{Design principles}

Three choices define an item. First, the removed part is \emph{functionally essential}: the object cannot do what it is for without it. A bicycle without its front wheel, an aircraft without its vertical stabiliser, a mug without its handle, a phone without its charging port, a cup lid without its sip hole are all objects that no longer work, and a model that knows the object knows the part should be there. The absence is therefore in tension with the prior, which is the condition the benchmark is designed to measure.

Second, exactly one part is removed per item, so each question has a single correct answer and the difficulty of an item is attributable to one part. The 452 items cover 387 distinct part names; the most frequent (front wheel, handle, rear wheel, handlebar, camera) each appear at most seven times (Table~\ref{tab:stats}).

Third, the unedited photograph is the per-item control. Every model answers the same question on the original and on the edit, so the original establishes that the model recognises the part, sees it at that viewpoint and resolution, and is not pushed towards ``not visible'' by the wording (Section~\ref{sec:protocol}).

The items follow a within-photograph design. The 452 items come from 175 source photographs, and 98 of these photographs contribute several items, up to ten, each with a different part removed, so that lighting, viewpoint, background and object identity are shared across the items of one photograph.

\subsection{Construction}

Photographs were selected by hand so that the object is an ordinary one with a recognisable function and the target part is visible and unoccluded in the original. The part was removed with the generative editor Nano Banana~2 for 64\% of items and with a manual eraser (cleanup.pictures) for 36\%. All source photographs are from Wikimedia Commons; every image is documented with its URL and licence in the datasheet, and each image and its edit are released under the licence of the source photograph.

Edited images are produced by the editor at its own output resolution, which differs from the original's for most items (for example a $450\times450$ original and a $1024\times1024$ edit). Every image, original or edited, passes through the same preprocessing before a model sees it (RGB, longest side at most 1600~px, JPEG quality~95), and each model call carries a single image.

\begin{table}[t]
    \centering
    \small
    \begin{tabular}{lr}
        \toprule
        Items (original + edited pairs) & 452 \\
        Distinct part names & 387 \\
        Source photographs (all from Wikimedia Commons) & 175 \\
        Photographs contributing several items (up to 10) & 98 \\
        Items missed by all nine models & 73 \\
        \midrule
        Edited with Nano Banana~2 / manual eraser & 64\% / 36\% \\
        \midrule
        Preprocessing for every call & longest side $\leq$ 1600~px, JPEG 95 \\
        \bottomrule
    \end{tabular}
    \caption{\bench{} at a glance.}
    \label{tab:stats}
\end{table}

\subsection{Human verification}

Three annotators each saw all 452 edited and all 452 original images in a shuffled order, without being told which were edited, and answered the three-way question put to the models together with an editing-artifact score from 1 to 5. A fixed rule returned an item for re-editing when at least two annotators answered \vis{} or \hs{} on its edited image, at least two gave an artifact score of 4 or more, or at least two answered \nv{} on its original; 21 items were returned at least once, and all 452 released items pass this rule, so by construction the annotators' majority is \nv{} on every released edited image and \vis{} or \hs{} on every released original. Pooled over the individual answers ($3\times452$ per split), the annotators answer \nv{} on 97.6\% of edited images (\vis{} 0.0\%, \hs{} 2.4\%) and answer the originals correctly on 100.0\% (false \nv{} 0.0\%). Agreement on edited images is Gwet's AC1 $= 0.95$ \citep{gwet_ac1_2008}; the mean artifact score is 2.8 on edited and 1.4 on original images; the mean answer time is 4.3~s. Protocol and agreement statistics are in Appendix~\ref{app:human}.

\subsection{Release and datasheet}

The release contains the images, a \texttt{metadata.csv} with one row per item (id, part, source, licence, editor, source-photograph id), every per-item model output in JSONL, the verbatim prompts and the evaluation scripts. A datasheet in the format of \citet{gebru_datasheets_2021}, covering motivation, composition, collection, licensing and intended use, accompanies the human-verification protocol in Appendix~\ref{app:human}.

\section{Evaluation Protocol}
\label{sec:protocol}

\paragraph{The question.}
Every model receives one image and one question: ``Is there a/an $\langle$part$\rangle$ visible in the image'', followed by three options: 0, ``It is clearly visible'' (\vis{}); 1, ``It is visible but hard to see'' (\hs{}); 2, ``It is not visible'' (\nv{}). We call this the three-way question. The prompt asks for a reasoning before the answer and for a JSON object whose first field is a brief visual justification and whose second is the integer; the verbatim text is in Appendix~\ref{app:prompts}. The same question, options and order are used for the originals, the human annotators and the open-weight model.

\paragraph{Scoring.}
On an edited image the correct answer is \nv{}; \vis{} and \hs{} are both wrong, because each asserts a part that is not in the picture. On an original the correct answers are \vis{} and \hs{}, and \nv{} is a false absence. The \nv{} rate on edited images is therefore the accuracy on the task, and we report it together with the rate of correct answers on originals and the balanced accuracy, the mean of the two.

\paragraph{Models and calls.}
Nine proprietary and open-weight VLMs released in 2026 are queried through OpenRouter on 2026-10-03 and 2026-10-04 with the provider-default temperature and no system prompt (identifiers and settings in Appendix~\ref{app:models}). Every image is converted to RGB, its longest side reduced to at most 1600\,px, encoded as JPEG at quality 95 and sent as a base64 data URL. Each request carries one item; the original and the edited copy are never shown together. A call is attempted up to five times with exponential backoff, and items without a recoverable answer are re-queued in later rounds until every item has one. The parser strips thinking blocks and code fences, parses the full text, then the last JSON object, then an answer field stated in prose.

\paragraph{Statistics.}
Every rate carries a Wilson 95\% confidence interval. Comparisons between two conditions on the same items (a second run, or a report delivered before or after the answer) use the exact McNemar test on the paired answers \citep{mcnemar_1947}; comparisons across the nine models use a sign test. A comparison of items from single-item and multi-item photographs is in Appendix~\ref{app:extra}.

\paragraph{Detector-report settings.}
A fixed, simulated detector report states that the part was not found:
\begin{quote}\ttfamily\small
External object detector result:\\
Target object part: \{part\}\\
Detection result: NOT DETECTED\\
Detection confidence: 0.03\\
The detector did not find the specified target object part.
\end{quote}
In the \emph{pre} setting the report is placed inside the single user turn, before the question, and the model is asked to use both the image and the report. In the \emph{post} setting the model first answers the direct query; a second, independent single-turn request then repeats the image, quotes the complete first response verbatim, adds the report, and asks the model to reconsider its answer using the image and this additional detector evidence. The two settings deliver the same report; the post request differs in that it quotes an earlier answer and asks the model to reconsider it, so the post--pre difference includes the effect of that request. The first request of the post run is an independent draw of the direct query under the same prompt, and Section~\ref{sec:results} uses it to measure run-to-run agreement.

\paragraph{Cost.}
One direct-query pass over the nine models costs about 18 USD; the full set of API runs in this paper costs about 150 USD.

\section{Results}
\label{sec:results}

\subsection{Absence detection}
\label{sec:main}

Nine deployed VLMs answer \nv{} on 39.6--70.1\% of edited images and answer the originals correctly on 94.2--99.8\%. Table~\ref{tab:main} reports the direct query of Section~\ref{sec:protocol} on all 452 items per model, with Wilson 95\% confidence intervals on both accuracies. The edited \nv{} rate has a mean of 50.6\%: five of the nine models are below 50\% (GPT-6 Luna 42.7, GPT-6.1 Sol 46.9, GLM 5.3 FlashX 46.2, Kimi K2.5 45.1, MiMo v2.6 Pro 39.6), Grok 4.7 is at 50.0, and the remaining three reach 54.2 (Claude Sonnet 5.5), 60.4 (Qwen3.8 Omni Flash) and 70.1 (Gemini 3.8 Flash). Balanced accuracy runs from 67.9 to 83.6. The three annotators reach 97.6\% on edited images and 100.0\% on originals pooled over their individual answers (balanced 98.8\%; Section~\ref{sec:benchmark}).

\begin{table}[t]
    \centering
    \small
    \resizebox{\linewidth}{!}{\begin{tabular}{lccccc}
        \toprule
        Model & Originals correct \% [CI] & Edited \vis{} \% & Edited \hs{} \% & Edited \nv{} \% [CI] & Balanced acc.\ \% \\
        \midrule
        Claude Sonnet 5.5   & 98.5 [96.8, 99.2]  & 23.5 & 22.3 & 54.2 [49.6, 58.7] & 76.3 \\
        Gemini 3.8 Flash    & 97.1 [95.1, 98.3]  & 20.6 & 9.3  & 70.1 [65.8, 74.2] & 83.6 \\
        GPT-6 Luna          & 99.1 [97.7, 99.7]  & 39.2 & 18.1 & 42.7 [38.2, 47.3] & 70.9 \\
        GPT-6.1 Sol         & 99.8 [98.8, 100.0] & 35.6 & 17.5 & 46.9 [42.3, 51.5] & 73.3 \\
        Grok 4.7            & 94.2 [91.7, 96.0]  & 33.2 & 16.8 & 50.0 [45.4, 54.6] & 72.1 \\
        GLM 5.3 FlashX      & 97.8 [96.0, 98.8]  & 34.3 & 19.5 & 46.2 [41.7, 50.8] & 72.0 \\
        Kimi K2.5           & 96.5 [94.3, 97.8]  & 50.2 & 4.6  & 45.1 [40.6, 49.7] & 70.8 \\
        MiMo v2.6 Pro       & 96.2 [94.1, 97.6]  & 57.1 & 3.3  & 39.6 [35.2, 44.2] & 67.9 \\
        Qwen3.8 Omni Flash  & 96.5 [94.3, 97.8]  & 26.5 & 13.1 & 60.4 [55.8, 64.8] & 78.4 \\
        \midrule
        Human annotators (3, individual answers) & 100.0 & 0.0 & 2.4 & 97.6 & 98.8 \\
        \bottomrule
    \end{tabular}}
    \caption{Direct query on \bench{}, $n=452$ edited and 452 original images per model. On an edited image \nv{} is correct; on an original \vis{} or \hs{} is correct. Balanced accuracy is the mean of the two accuracies. Brackets give Wilson 95\% confidence intervals. The last row pools the individual answers of three annotators under the same protocol ($3\times452$ per split); by construction of the verification rule, their majority is \nv{} on every released edited item (Section~\ref{sec:benchmark}).}
    \label{tab:main}
\end{table}

The wrong answers more often assert that the absent part is clearly visible than that it is hard to see. On edited images \vis{} exceeds \hs{} on every model, from 23.5 against 22.3 for Claude Sonnet 5.5 to 57.1 against 3.3 for MiMo v2.6 Pro; the \vis{} rate spans 20.6--57.1\%.

Every model answers 94.2--99.8\% of the originals correctly, with false \nv{} on 0.2--5.8\%, so the question wording, the image and the model's knowledge of the part name support a correct answer when the part is present.

\begin{figure}[t]
    \centering
    \includegraphics[width=0.62\linewidth]{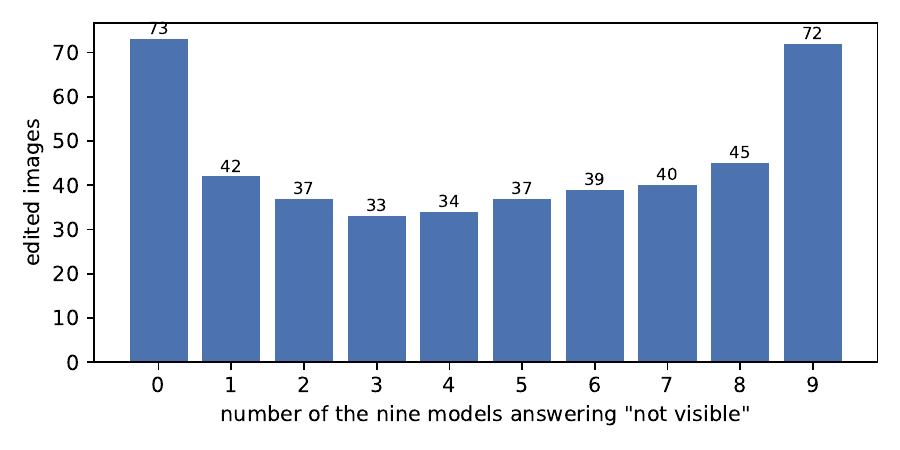}
    \caption{Item difficulty on the 452 edited images: the number of items on which exactly $k$ of the nine models answer \nv{}. 73 items are missed by every model and 72 are solved by all nine.}
    \label{fig:difficulty}
\end{figure}

Item difficulty spans the full range, with a concentration at both ends. Figure~\ref{fig:difficulty} counts, for each item, how many of the nine models answer \nv{}: 73 items (16\%) are missed by every model and 72 are solved by all nine, with 33--45 items at each intermediate count. As on every released item, the annotators' majority answer on these 73 is \nv{}.

We measure run-to-run variation with the direct run and the independent first request of the post run of Section~\ref{sec:evidence}, which use the same prompt under the provider-default temperature. The two runs agree on \nv{}-versus-not on 77.4--94.5\% of items per model (three-way agreement 68.4--93.6\%), and each model's \nv{} rate differs between the runs by $-2.9$ to $+3.9$ points (exact McNemar $p = 0.029$ for Claude Sonnet 5.5 and $p > 0.1$ for every other model; Table~\ref{tab:apprun}). Sampling changes the answer on individual items, but no model's rate moves by more than 4 points.

\subsection{The same report, before and after the answer}
\label{sec:evidence}

\begin{figure}[t]
    \centering
    \includegraphics[width=0.9\linewidth]{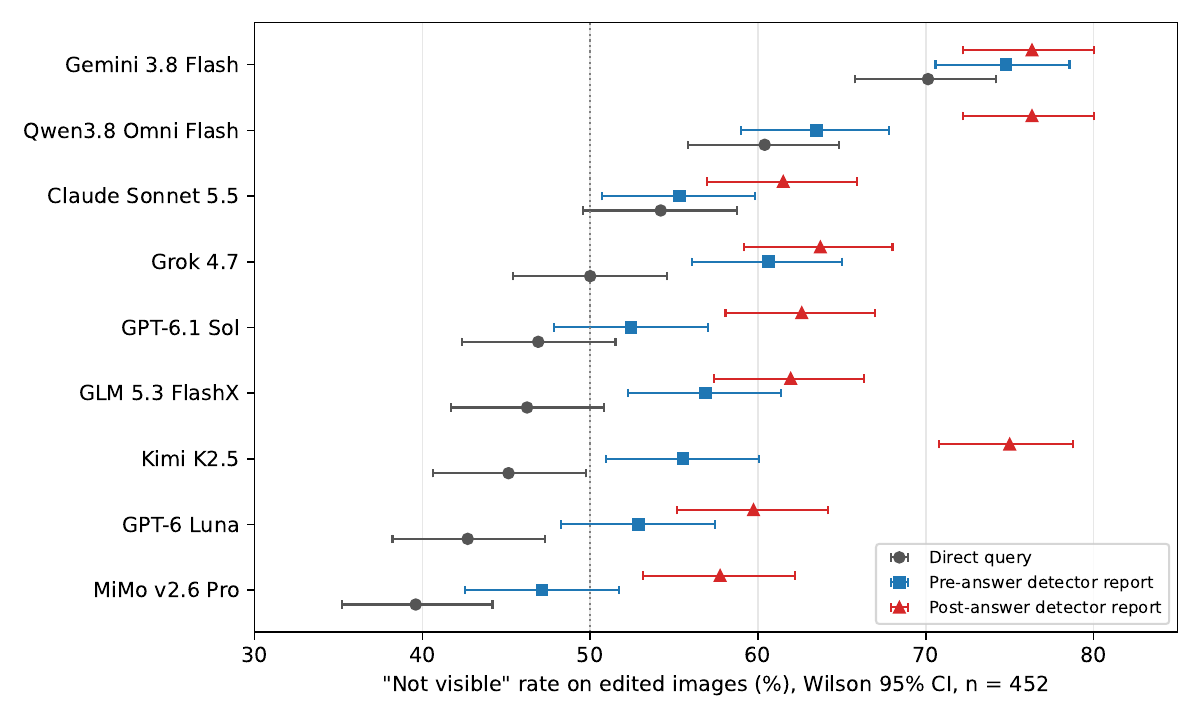}
    \caption{\nv{} rate on edited images per model under the direct query, the report delivered before the question (pre) and the report delivered after the answer (post). Post exceeds pre on 9 of 9 models.}
    \label{fig:injection}
\end{figure}

The same detector report raises the \nv{} rate on every model, and raises it more when delivered after the model has answered than when delivered before the question. The report (Section~\ref{sec:protocol}) names the target part and states that it was not detected, with confidence 0.03, and is identical in both settings: \emph{pre} places it inside the single user turn before the question; \emph{post} is a second, independent single-turn request that repeats the image, quotes the complete first response verbatim, adds the report and asks the model to reconsider using the image and the detector evidence. Table~\ref{tab:evidence} and Figure~\ref{fig:injection} report the \nv{} rate under the direct query and both settings, $n=452$ edited images per model.

\begin{table}[t]
    \centering
    \small
    \resizebox{\linewidth}{!}{\begin{tabular}{lccccccc}
        \toprule
        Model & Direct \nv{} & Pre \nv{} & Post \nv{} & $\Delta$ post$-$direct ($p$) & wrong$\rightarrow$\nv{} & \nv{}$\rightarrow$wrong & Errors kept \% \\
        \midrule
        Claude Sonnet 5.5   & 54.2 & 55.3 & 61.5 & $+7.3$ ($<0.001$)  & 30  & 10 & 84.5 \\
        Gemini 3.8 Flash    & 70.1 & 74.8 & 76.3 & $+6.2$ ($<0.001$)  & 25  & 1  & 80.9 \\
        GPT-6 Luna          & 42.7 & 52.9 & 59.7 & $+17.0$ ($<0.001$) & 75  & 3  & 70.5 \\
        GPT-6.1 Sol         & 46.9 & 52.4 & 62.6 & $+15.7$ ($<0.001$) & 80  & 0  & 67.9 \\
        Grok 4.7            & 50.0 & 60.6 & 63.7 & $+13.7$ ($<0.001$) & 75  & 0  & 68.6 \\
        GLM 5.3 FlashX      & 46.2 & 56.9 & 61.9 & $+15.7$ ($<0.001$) & 72  & 0  & 70.5 \\
        Kimi K2.5           & 45.1 & 55.5 & 75.0 & $+29.9$ ($<0.001$) & 140 & 5  & 43.5 \\
        MiMo v2.6 Pro$^{\dagger}$ & 39.6 & 47.1 & 57.7 & $+18.1$ ($<0.001$) & 65  & 4  & 73.7 \\
        Qwen3.8 Omni Flash  & 60.4 & 63.5 & 76.3 & $+15.9$ ($<0.001$) & 81  & 0  & 56.9 \\
        \bottomrule
    \end{tabular}}
    \caption{Detector report on edited images, $n=452$ per model. \nv{} rates (\%) under the direct query, the report delivered before the question (pre) and the report delivered after the model has answered (post); $p$ from exact McNemar against the direct run. The flip counts compare the first request of the post run with the post request; ``errors kept'' is the share of first-request wrong answers that remain wrong after the report. $^{\dagger}$For MiMo v2.6 Pro, 14 responses to the first request of the post run contain no parsable answer. The first response is quoted as returned and is not re-queued; the re-queuing of Section~\ref{sec:protocol} applies to the scored answer of each setting. The post request was therefore issued for all 452 items and the post \nv{} rate is over 452 (261 \nv{}), while the flip counts and errors kept are over the 438 items with a parsed first answer (191 \nv{}, 247 wrong, 182 of them kept). Full answer distributions are in Appendix~\ref{app:extra}.}
    \label{tab:evidence}
\end{table}

Delivered after the answer, the report raises \nv{} on all nine models by 6.2--29.9 points. The pre setting raises \nv{} by 1.1--10.6 points (mean 50.6 to 57.7), significant on seven of nine models (Claude Sonnet 5.5 $p=0.533$, Qwen3.8 Omni Flash $p=0.098$); the post setting raises it to a mean of 66.1, significant on all nine ($p<0.001$). Post exceeds pre on 9 of 9 models (sign test $p=0.004$), by 1.5--19.5 points; the largest change is Kimi K2.5, from 45.1 to 75.0.

Most flips are towards \nv{}. Between the first request of the post run and the post request, 25--140 answers per model move from a wrong answer to \nv{} and 0--10 move from \nv{} to a wrong answer; four models move no answer in the wrong direction.

The share of a model's first-request wrong answers that remain wrong after the post request is 43.5--84.5\%, from Kimi K2.5 to Claude Sonnet 5.5, although the image is in front of the model and the report states that the part was not found.

\subsection{Without naming the part}
\label{sec:framing}

An open-ended framing that never names the part gives the same ranking of models as the three-way question. The prompt asks the model to list every part of the object it can see; an item counts as detected when the removed part is absent from the list, judged by manual matching under a written synonym rule (the synonym rule is released with the code). The framing is run on five models that span the range of three-way \nv{} rates (39.6--70.1\%).\footnote{The open-ended framing was measured on the 464-item set from which the released 452 items were drawn, before the exclusion of 12 items; the three-way column of Table~\ref{tab:framing} is the direct query on the 452 items of Table~\ref{tab:main}.} A model may also omit a part that is present, so the originals supply the correction.

\begin{table}[t]
    \centering
    \small
    \resizebox{\linewidth}{!}{\begin{tabular}{lcccc}
        \toprule
        Model & Edited: removed part absent \% & Originals: present part listed \% & Corrected detection \% & Three-way \nv{} (direct, 452) \% \\
        \midrule
        Gemini 3.8 Flash  & 80.2 & 86.4 & 66.6 & 70.1 \\
        Claude Sonnet 5.5 & 72.8 & 89.0 & 61.9 & 54.2 \\
        GPT-6.1 Sol       & 63.1 & 88.5 & 51.6 & 46.9 \\
        Kimi K2.5         & 57.3 & 88.1 & 45.5 & 45.1 \\
        MiMo v2.6 Pro     & 50.6 & 89.6 & 40.2 & 39.6 \\
        \bottomrule
    \end{tabular}}
    \caption{Open-ended framing, $n=464$ edited and 464 original images per model (footnote in the text). The model lists the parts it sees without being told which part to look for. Corrected detection subtracts the omission rate on originals (100 minus the ``present part listed'' column) from the rate at which the removed part is absent from the list on edited images, computed from unrounded rates. The last column repeats the direct three-way \nv{} rate of Table~\ref{tab:main}.}
    \label{tab:framing}
\end{table}

Models omit a present part 10--14\% of the time and the removed part 51--80\% of the time. The present part is listed on 86.4--89.6\% of originals; the removed part is absent from the list on 50.6--80.2\% of edited images. Subtracting the omission rate gives a corrected detection of 40.2--66.6\%, which differs from the three-way \nv{} rate of the same model by 0.4--7.7 points (largest for Claude Sonnet 5.5, 61.9 against 54.2) and orders the five models identically: Gemini 3.8 Flash, Claude Sonnet 5.5, GPT-6.1 Sol, Kimi K2.5, MiMo v2.6 Pro.

\section{Perception or Readout?}
\label{sec:mechanism}

Two accounts are consistent with Section~\ref{sec:results}. Under the first, the absence never reaches the representation from which the answer is produced: the state describes the object the model expects, and the answer reports that description. Under the second, the representation registers the absence and the answer is produced by a step that weights a prior over the evidence. We use \emph{readout} for any procedure that turns a hidden state into a prediction: a trained probe, the logit lens, or the model's own answer. Two experiments bear on the two accounts. Open-vocabulary detectors show how well a text-queried visual match separates edited from original images. An open-weight VLM run with the same prompt shows, layer by layer, how well its state separates them, how well its own answers do, and what happens when a separating direction is added to the residual stream.

\subsection{Open-vocabulary detectors}
\label{sec:detectors}

A detector queried with the part name separates edited from original images at an AUROC of 0.64 or below. OWLv2-large \citep{minderer_scaling_2024}, YOLOE-26l \citep{wang_yoloe_2025} and Grounding DINO-base \citep{liu_groundingdino_2024} receive every original and edited image with the part name as the text query; the score is the maximum box confidence for that name, the original is the positive class, and confidence intervals are bootstrapped over items.

\begin{figure}[t]
    \centering
    \includegraphics[width=0.95\linewidth]{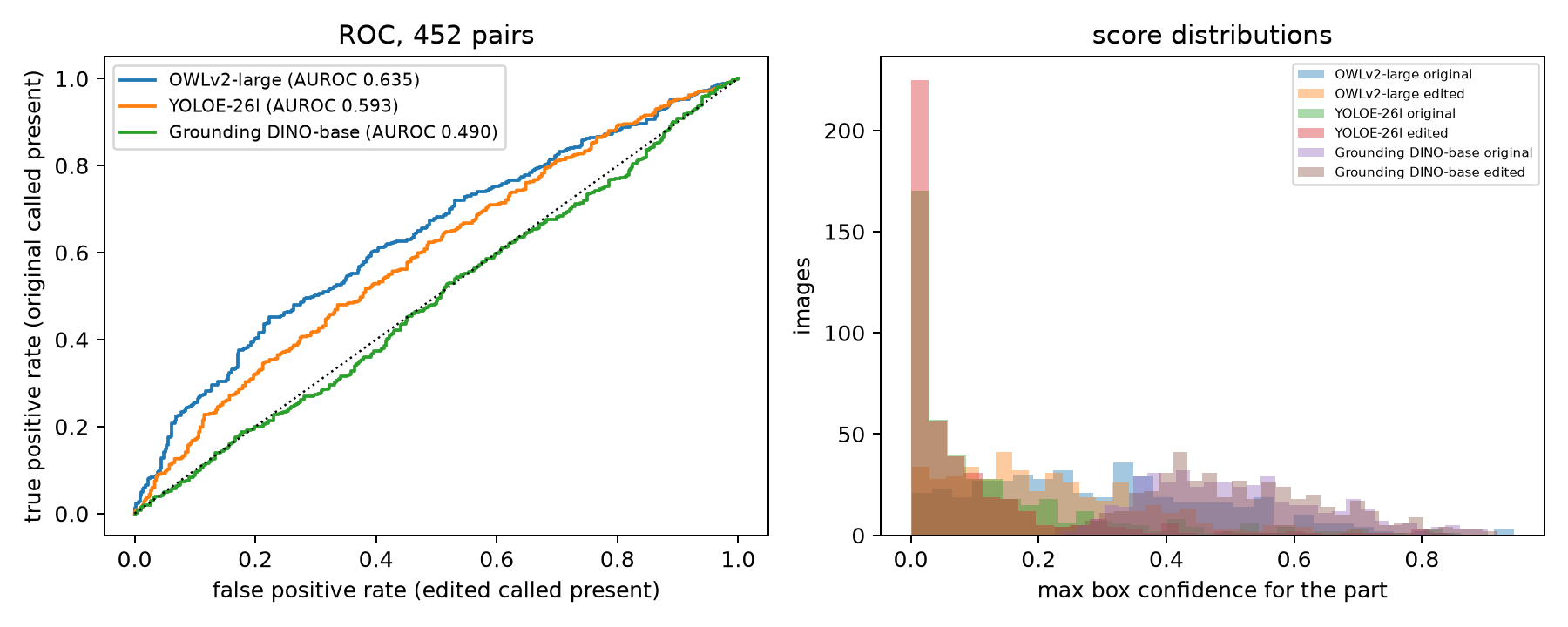}
    \caption{ROC curves (left) and score distributions on original and edited images (right) for the three detectors.}
    \label{fig:detectors}
\end{figure}

OWLv2 reaches an AUROC of 0.635 [0.615, 0.657], YOLOE 0.593 [0.572, 0.615] and Grounding DINO 0.490 [0.474, 0.505] (Table~\ref{tab:detectors}); the best balanced accuracy at any threshold is 61.4\%, 57.2\% and 50.9\%, and at the threshold that keeps 95\% of original parts it is 53.1\%, 52.9\% and 50.4\%. At their default thresholds OWLv2 and Grounding DINO err in the same direction as the VLMs: they still report the part on 74.1\% and 89.2\% of edited images (and on 85.8\% and 90.0\% of originals). YOLOE errs the other way, reporting the part on only 16.6\% of originals and 9.5\% of edited images. The original outscores its edited copy on 68.6\%, 57.5\% and 44.7\% of pairs, and Grounding DINO's scores on the two images of a pair are within 10\% of each other for 68\% of pairs. Figure~\ref{fig:detectors} shows the ROC curves and the overlap of the score distributions. None of the three detectors gives a usable absence signal on these pairs at any threshold.

\begin{table}[t]
    \centering
    \small
    \resizebox{\linewidth}{!}{\begin{tabular}{lccccc}
        \toprule
        Detector & AUROC [95\% CI] & AP & Best bal.\ acc.\ (\%) & Bal.\ acc.\ at 95\% orig.\ recall (\%) & Orig.\ recall / edited reject (\%) \\
        \midrule
        OWLv2-large & 0.635 [0.615, 0.657] & 0.641 & 61.4 ($t=0.324$) & 53.1 & 85.8 / 25.9 ($t=0.1$) \\
        YOLOE-26l & 0.593 [0.572, 0.615] & 0.582 & 57.2 ($t=0.065$) & 52.9 & 16.6 / 90.5 ($t=0.25$) \\
        Grounding DINO-base & 0.490 [0.474, 0.505] & 0.495 & 50.9 ($t=0.314$) & 50.4 & 90.0 / 10.8 ($t=0.35$) \\
        \bottomrule
    \end{tabular}}
    \caption{Open-vocabulary detectors on the 452 pairs. Score is the maximum box confidence for the part name; the original is the positive class. Best balanced accuracy is over all thresholds; the last column is at each detector's default threshold.}
    \label{tab:detectors}
\end{table}

\subsection{An open-weight model, layer by layer}
\label{sec:probes}

Qwen3-VL-8B-Instruct, run locally with the same prompt, shows the pattern of Table~\ref{tab:main} at a lower \nv{} rate than any of the nine API models: \nv{} on 29.4\% [25.4, 33.8] of edited images (\vis{} 69.2\%, \hs{} 1.3\%), originals correct on 97.1\%, balanced accuracy 63.3\%; greedy decoding gives the same pattern.

A linear probe \citep{alain_probes_2016}, PCA to 256 dimensions followed by logistic regression, is trained to separate edited from original images from the residual state of the last prompt token under the direct-query prompt, the position at which the model's response begins, at every layer of the language model, on 448 pairs from 173 photographs. Folds are grouped by source photograph so that no photograph contributes to both training and test, confidence intervals are cluster bootstraps over photographs, and a permutation null gives $0.50 \pm 0.02$.

\begin{figure}[t]
    \centering
    \includegraphics[width=0.85\linewidth]{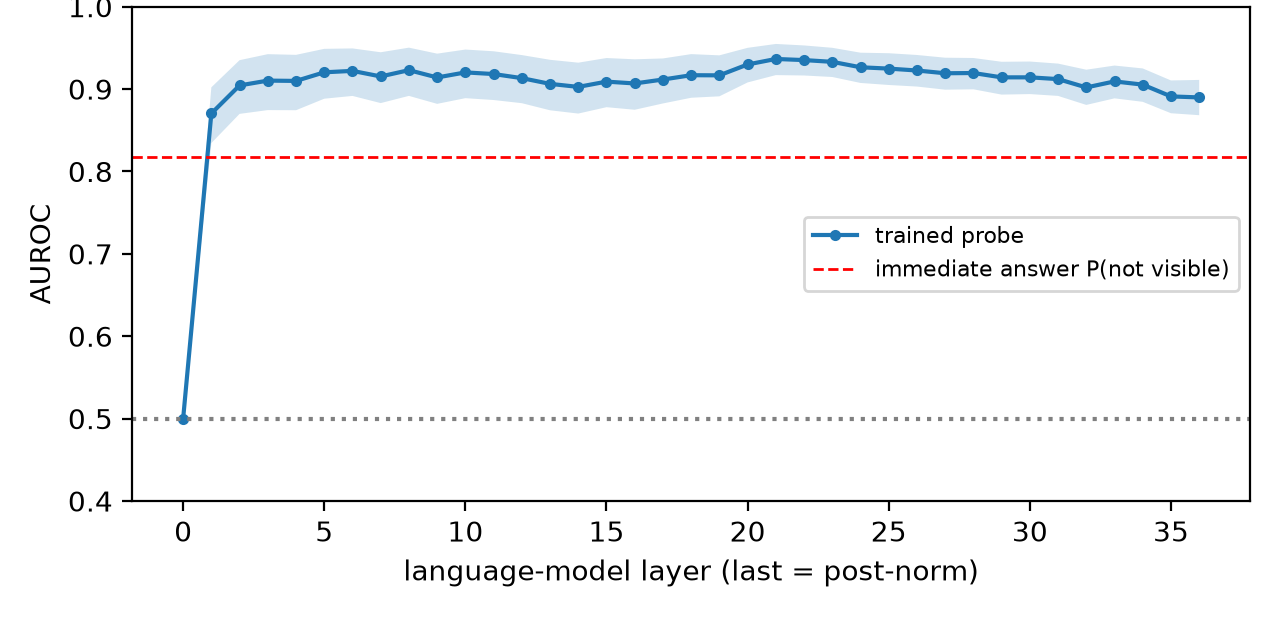}
    \caption{AUROC of the linear probe (edited versus original) by layer on the last prompt token of Qwen3-VL-8B-Instruct (band: cluster-bootstrap 95\% CI), with the AUROC of the model's own immediate answer as the dashed line. Per-layer values of the probe and the logit lens are listed in Tables~\ref{tab:applayers} and~\ref{tab:applens}.}
    \label{fig:probe}
\end{figure}

Edited and original images are linearly separable at every depth after the embeddings (Table~\ref{tab:probes}, Figure~\ref{fig:probe}). The probe is at chance on the embeddings, 0.87 at layer 1, and between 0.89 and 0.94 from layer 2 to layer 36: 0.917 [0.889, 0.942] at mid-depth, 0.919 [0.899, 0.938] at three-quarter depth and 0.890 [0.868, 0.911] at the final post-norm state. A logit lens \citep{nostalgebraist_logitlens_2020}, the probability of the answer digit 2 (``not visible'') among the three answer digits read from each layer of the same state through the final norm and the unembedding without training, separates the two sets at 0.29--0.64 through layer 32 and at 0.65--0.75 over layers 33--36 (Table~\ref{tab:applens}). The model's \emph{immediate answer} is its distribution over the three answer digits when the request for reasoning is removed and the reply is prefilled up to the digit (prompt in Appendix~\ref{app:prompts}); averaged over both option orders, its probability of ``not visible'' separates edited from original images at 0.817 [0.792, 0.843], with a mean of 0.42 on edited images against 0.09 on originals. The answer after reasoning is \nv{} on 29.4\% of edited images. The state thus separates the two sets better than either of the model's own answers does, from the first layer to the last. The probe label is edited versus original, and an edit leaves traces of its own, so this separability does not by itself show that the state encodes the absence of the part (Section~\ref{sec:discussion}).

\begin{table}[t]
    \centering
    \small
    \resizebox{\linewidth}{!}{\begin{tabular}{llcl}
        \toprule
        Readout & Depth & AUROC [95\% CI] & Operating point \\
        \midrule
        Linear probe & embeddings (layer 0) & 0.50 & --- \\
        Linear probe & layer 1 & 0.87 & --- \\
        Linear probe & mid-depth (layer 18) & 0.917 [0.889, 0.942] & --- \\
        Linear probe & three-quarter depth (layer 27) & 0.919 [0.899, 0.938] & --- \\
        Linear probe & final post-norm state (layer 36) & 0.890 [0.868, 0.911] & --- \\
        Logit lens & layers 1--32 & 0.29--0.64 & --- \\
        Logit lens & layers 33--36 & 0.65--0.75 (0.75 at output) & --- \\
        Immediate answer (no chain of thought) & output & 0.817 [0.792, 0.843] & $P(\nv{})$ 0.42 edited, 0.09 original \\
        Answer after chain of thought & output & --- & \nv{} 29.4\% edited, 97.1\% originals correct \\
        \bottomrule
    \end{tabular}}
    \caption{Separation of edited from original images by each readout of Qwen3-VL-8B-Instruct. AUROC rows: 448 pairs from 173 photographs, folds grouped by photograph, permutation null $0.50 \pm 0.02$. The last row is the direct query on all 452 pairs (133 of 452 edited images answered \nv{}).}
    \label{tab:probes}
\end{table}

Steering along a separating direction changes the answers (Table~\ref{tab:steer}). Directions are fitted on training photographs and evaluated on 179 held-out pairs; the probe direction separates the held-out pairs at 0.90 by projection alone. Added at the output of layer 18 with magnitudes up to 27\% of the state norm, the probe direction raises \nv{} in the immediate answer from 37.4\% to 54.7\% on edited images and from 5.6\% to 13.4\% on originals; at layers 27 and 33 it has little effect. The AUROC of the immediate answer stays at 0.825--0.829 throughout, which is expected when the same vector is added to every image and therefore says little about the readout. With chain of thought, the class-mean direction at layer 27 raises \nv{} on edited images from 27.9\% to 45.8\% and 65.2\% at the two larger magnitudes while false-\nv{} on originals rises from 2.8\% to 7.8\% and 21.8\%, so the balanced accuracy of the reasoned answer rises from 62.6\% to 69.0\% and 71.7\%. The probe direction at the same layer leaves the reasoned answer unchanged (Appendix~\ref{app:extra}).

\begin{table}[t]
    \centering
    \small
    \resizebox{\linewidth}{!}{\begin{tabular}{llllccc}
        \toprule
        Direction & Layer & Answer & Magnitude $\alpha$ & Edited \nv{} (\%) & Original false-\nv{} (\%) & Balanced acc.\ (\%) \\
        \midrule
        Probe & 18 & immediate & 0 $\to$ 32 & 37.4 $\to$ 54.7 & 5.6 $\to$ 13.4 & 65.9 $\to$ 70.7 \\
        Probe & 27, 33 & immediate & 0 $\to$ 32 & 37.4 $\to$ 37.4, 39.7 & 5.6 $\to$ 5.0, 5.6 & 65.9 $\to$ 66.2, 67.0 \\
        Class mean & 27 & chain of thought & 0 $\to$ 4 & 27.9 $\to$ 45.8 & 2.8 $\to$ 7.8 & 62.6 $\to$ 69.0 \\
        Class mean & 27 & chain of thought & 0 $\to$ 8 & 27.9 $\to$ 65.2 & 2.8 $\to$ 21.8 & 62.6 $\to$ 71.7 \\
        \bottomrule
    \end{tabular}}
    \caption{Activation steering on 179 held-out pairs. The direction is added at the layer output; $\alpha$ is in units of the class-mean gap along the direction, and the largest value for the probe direction corresponds to 27\% of the state norm. The immediate-answer AUROC is 0.825--0.829 in all probe-direction rows. The full grid is in Appendix~\ref{app:extra}.}
    \label{tab:steer}
\end{table}

The wording of the prompt also changes the answer more than it changes the probe. On the same model and images, the direct-query prompt yields \nv{} on 29.4\% of edited images; a system message, ``You are a careful visual inspector'', with a step-by-step instruction yields 45.6\% (false-\nv{} 7.5\%). Under the two prompts the probe reads 0.89--0.94 and 0.92--0.96 from layer 2 on.

Taken together, in this 8B model the hidden state separates edited from original images at 0.89--0.94 at every depth, the immediate answer separates them at 0.82, and the answer after reasoning is correct on 29\% of edited images; steering and a change of wording both move the answer by 16 points or more. This pattern is consistent with the second account, a loss between the representation and the answer, but it does not rule out the first, because the probe may separate the images by traces of the edit. Whether the pattern holds for the API models, which we could not inspect, is open.

\section{Discussion}
\label{sec:discussion}

\paragraph{What the benchmark measures.}
\bench{} measures one quantity: whether a model reports that an essential part is absent from a real photograph when it is. All nine models fall well short of the annotators on this quantity while answering the originals almost perfectly, and the errors are concentrated: 73 items are missed by all nine.

\paragraph{Implications for inspection deployments.}
On edited images, a correct report delivered after the model has answered raises the ``not visible'' rate on every one of the nine models, by 6 to 30 points, and reverses few correct answers: between 0 and 10 answers per model move from ``not visible'' to wrong, against 25 to 140 that move from wrong to ``not visible''. Because 44--85\% of a model's first-request errors remain after the report, a deployment that relies on a VLM to confirm the presence of a safety-critical part would still need an independent check. These figures are for edited images, where the report is true. Industrial anomaly datasets \citep{bergmann_mvtec_2019} and multimodal inspection benchmarks \citep{jiang_mmad_2024} describe this setting for manufactured products under controlled imaging; \bench{} adds ordinary objects in real photographs, with the originals as controls.

\paragraph{Implications for model design.}
In the open-weight model the hidden state separates edited from original images better than the answer does, and steering along the class-mean direction recovers part of the gap in the reasoned answer (balanced accuracy 62.6\% to 71.7\%). If the separable signal reflects the absence of the part and the pattern holds for larger models, interventions on how the answer is produced from the state, through training objectives or steering, may help more than a stronger visual encoder. This would agree with reports that VLMs attend to or encode the relevant evidence on many of the questions they answer wrongly \citep{seeing_not_believing,encoded_but_unused}, and with the competition between a prior and contrary evidence documented for text in work on knowledge conflict \citep{wu_clasheval_2025, jin_tug--war_2024} and sycophancy \citep{sharma_sycophancy_2024}.

\paragraph{Limitations.}
The probe of Section~\ref{sec:probes} is trained on edited versus original images, and 64\% of the edits were re-rendered by a generative editor and resized. The probe reaches 0.87 at the first layer, and on the visual tokens a probe on a control box far from the part separates the images as well as one on the removed-part box (0.734 against 0.727), so the separable signal may be the edit and not the absence. The representation analysis covers one 8B model, and we have no evidence on the nine API models. The detector report was delivered only on edited images, where it is true, and the post request also asks the model to reconsider an answer it is shown, so the post--pre difference cannot be attributed to timing alone. The human rates come from the annotators whose answers decided which items were re-edited, and are an upper reference. Every prompt is in English with a single wording, and each item removes one part.

\paragraph{Broader impact.}
The benchmark is intended to make VLM-based inspection safer by measuring a failure that matters there. No image contains personal data. All images come from Wikimedia Commons and are documented with URL and licence in the datasheet; each edit is released under the licence of its source photograph and beside its original, so that every edit is auditable.

\paragraph{Reproducibility.}
The release contains the images, the metadata, the verbatim prompts (Appendix~\ref{app:prompts}), the evaluation code and every per-item model output as JSONL; the exact model ids, call dates and decoding settings are in Appendix~\ref{app:models}.

\section{Conclusion}
\label{sec:conclusion}

\bench{} asks whether a model can tell that an essential part is missing from a real photograph. Nine deployed VLMs answer correctly on 40--70\% of edited images, against 98\% for human annotators, and a detector report stating that the part was not found closes only part of the gap. In one open-weight model, edited and original images remain separable in the hidden state at every depth, which is consistent with an error that arises when the answer is produced but does not yet separate the absence of the part from traces of the edit. Probes with controlled contrasts and the same analysis on larger models are the next steps.

\bibliography{main}
\bibliographystyle{tmlr}

\appendix

\section{Models and settings}
\label{app:models}

Table~\ref{tab:appmodels} lists the nine deployed models with the OpenRouter identifiers under which they were called. All calls were made on 2026-10-03 and 2026-10-04 with the provider-default temperature, without a system prompt, and with the instruction to return a JSON object whose reasoning field precedes the answer field (Appendix~\ref{app:prompts}).

\begin{table}[h]
\centering
\small
\begin{tabular}{lll}
\toprule
Model & OpenRouter id & Calls \\
\midrule
Claude Sonnet 5.5 & \texttt{anthropic/claude-sonnet-5.5} & 2026-10-03/04 \\
Gemini 3.8 Flash & \texttt{google/gemini-3.8-flash} & 2026-10-03/04 \\
GPT-6 Luna & \texttt{openai/gpt-6-luna} & 2026-10-03/04 \\
GPT-6.1 Sol & \texttt{openai/gpt-6.1-sol} & 2026-10-03/04 \\
Grok 4.7 & \texttt{x-ai/grok-4.7} & 2026-10-03/04 \\
GLM 5.3 FlashX & \texttt{z-ai/glm-5.3-flashx} & 2026-10-03/04 \\
Kimi K2.5 & \texttt{moonshotai/kimi-k2.5} & 2026-10-03/04 \\
MiMo v2.6 Pro & \texttt{xiaomi/mimo-v2.6-pro} & 2026-10-03/04 \\
Qwen3.8 Omni Flash & \texttt{qwen/qwen3.8-omni-flash} & 2026-10-03/04 \\
\bottomrule
\end{tabular}
\caption{The nine API models, their OpenRouter identifiers and call dates. Shared settings: provider-default temperature, no system prompt, JSON answer with the reasoning field first, one image per call.}
\label{tab:appmodels}
\end{table}

\paragraph{Image preprocessing.}
Every image sent to an API is converted to RGB, its longest side is reduced to at most 1600~px, and it is encoded as JPEG at quality 95 and transmitted as a base64 data URL. The same preprocessing is applied to originals and edited images.

\paragraph{Transport.}
Each call is attempted up to five times with exponential backoff ($2^k$~s plus jitter, doubled on HTTP~429 and capped at 60~s). HTTP 400, 401, 402, 403 and 404 responses are not retried. Items without a recoverable answer are re-queued in later rounds until every item has an answer.

\paragraph{Parser.}
The parser strips \texttt{\textless{}think\textgreater{}} blocks and code fences, attempts to parse the whole text as JSON, then the last JSON object in the text, and finally an answer field stated in prose. The answer integer (0, 1 or 2) is mapped to \vis{}, \hs{} and \nv{}.

\paragraph{Local open-weight model.}
The representation analysis uses Qwen3-VL-8B-Instruct (\texttt{Qwen/Qwen3-VL-8B-Instruct}) run locally with transformers 5.20.1 in bf16. The prompt is identical to the API prompt. Decoding uses the checkpoint default sampling with $T = 0.7$; greedy decoding gives the same pattern. For the three-way answers each image receives the same preprocessing as in the API calls. For the probe features, the edited image is first brought into its original's frame so that both images of a pair occupy the same token grid, and the same preprocessing follows. Probes are trained on 448 pairs (four pairs lack a pixel-difference region) from 173 photographs, with folds grouped by source photograph and cluster-bootstrap confidence intervals.

\paragraph{Detector checkpoints.}
The three open-vocabulary detectors are OWLv2 (\texttt{owlv2-large-patch14-ensemble}), YOLOE (\texttt{yoloe-26l-seg}) and Grounding DINO (\texttt{grounding-dino-base}). The part name is used as the text query, and the score of an image is the maximum box confidence returned for that query.

\section{Prompts}
\label{app:prompts}

The prompts below are reproduced verbatim. \texttt{\{part\}} is the part name, \texttt{\{a/an\}} the article chosen for it, \texttt{\{options\}} the three numbered options of the direct query, \texttt{\{report\}} the detector report, and \texttt{\{...\}} the same JSON template as in the direct query.

\begin{promptbox}{{Three-way question (direct query, all nine models and the local model)}}
Is there \{a/an\} \{part\} visible in the image, give a reasoning first before answer the int\\
0. It is clearly visible\\
1. It is visible but hard to see\\
2. It is not visible\\[1ex]
Return JSON only, with the reasoning field first:\\
\{"justification": "brief visual justification", "answer": 0\}
\end{promptbox}

\begin{promptbox}{{Detector report (fixed text, inserted as \{report\})}}
External object detector result:\\
Target object part: \{part\}\\
Detection result: NOT DETECTED\\
Detection confidence: 0.03\\
The detector did not find the specified target object part.
\end{promptbox}

\begin{promptbox}{Pre-answer request (report inside the single user turn)}
Target object part:\\
\{part\}\\[1ex]
Before answering, you are provided with this external detector result:\\
\{report\}\\
Using both the image and the detector result, is there \{a/an\} \{part\} visible in the image? Give a reasoning first before answer the int\\
\{options\}\\[1ex]
Return JSON only, with the reasoning field first:\\
\{...\}
\end{promptbox}

\begin{promptbox}{{Post-answer request (second, independent single-turn request with the image)}}
We previously asked you to determine whether the target object part (\{part\}) is visible.\\[1ex]
Your previous answer was:\\
\{first response\}\\[1ex]
You are now provided with an external detector result:\\
\{report\}\\
Please reconsider your answer using the image and this additional detector evidence.\\
\{options\}\\[1ex]
Return JSON only, with the reasoning field first:\\
\{...\}
\end{promptbox}

\begin{promptbox}{Immediate answer without chain of thought (local model; assistant turn prefilled)}
Is there \{a/an\} \{part\} visible in the image, answer the int\\
\{options\}\\[1ex]
Return JSON only:\\
\{"answer": 0\}\\[1ex]
(assistant turn prefilled with) \{"answer":
\end{promptbox}

\begin{promptbox}{{System-message variant (local model, comparison in Section~\ref{sec:mechanism})}}
SYSTEM: You are a careful visual inspector. Answer strictly in the requested JSON format.\\[1ex]
USER: Is the \{part\} visible in this image?\\
Think step by step first, then answer with one option:\\
0 = it is clearly visible\\
1 = it is visible but hard to see\\
2 = it is not visible\\
Respond with JSON only: \{...\}
\end{promptbox}

\begin{promptbox}{Open-ended framing (names no part)}
Look at the object in this image and list every part of it that you can see.\\
Name each part on its own line, using the common name for the part.\\
Do not list parts that are not visible in the image.\\[1ex]
Return JSON only:\\
\{"parts": ["part 1", "part 2", "..."]\}
\end{promptbox}

\section{Human verification and datasheet}
\label{app:human}

\paragraph{Annotator instructions.}
Three annotators each saw all 452 edited and all 452 original images in a shuffled order, without being told which images were edited. The instruction text read: ``You will see one image at a time together with the name of one part of the object. Answer the question `Is there a \{part\} visible in the image?' with one of three options: it is clearly visible; it is visible but hard to see; it is not visible. Then rate how strongly the image shows signs of digital editing on a scale from 1 (no sign of editing) to 5 (obvious editing). Some images have been edited and some have not; judge each image on what you see.'' The three-way question is therefore the same question the models answer, and the artifact score is collected on every image.

\paragraph{Re-edit rule.}
A fixed rule sent an item back for re-editing when at least two annotators answered \vis{} or \hs{} on its edited image, when at least two gave the edited image an artifact score of 4 or higher, or when at least two answered \nv{} on its original. Every item in the released set of 452 passes this rule. Table~\ref{tab:apphuman} reports the agreement statistics.

\begin{table}[h]
\centering
\small
\begin{tabular}{lr}
\toprule
Statistic & Value \\
\midrule
Edited images, individual answers \nv{} (\%) & 97.6 \\
Edited images, individual answers \vis{} (\%) & 0.0 \\
Edited images, individual answers \hs{} (\%) & 2.4 \\
Originals, individual answers correct (\%) & 100.0 \\
Originals, individual answers false \nv{} (\%) & 0.0 \\
Items returned for re-editing during construction & 21 \\
Gwet's AC1 on edited images & 0.95 \\
Mean artifact score, edited images & 2.8 \\
Mean artifact score, original images & 1.4 \\
Mean answer time (s) & 4.3 \\
\bottomrule
\end{tabular}
\caption{Human verification on the 452 edited and 452 original images, three annotators who were not told which images were edited. Rates pool the individual answers ($3\times452$ per split); by construction of the re-edit rule, the majority vote is \nv{} on every released edited image and \vis{} or \hs{} on every released original. Agreement is Gwet's AC1 \citep{gwet_ac1_2008}.}
\label{tab:apphuman}
\end{table}

\paragraph{Datasheet.}
Following \citet{gebru_datasheets_2021}, the released datasheet documents the following. \emph{Motivation:} the benchmark measures whether a vision-language model can report that a functionally essential part of an object is absent from a real photograph, a question that inspection settings ask and that existing benchmarks of present content do not. \emph{Composition:} 452 items, each consisting of one real photograph and one edited copy in which exactly one essential part has been removed; 387 distinct part names; 175 source photographs, of which 98 contribute more than one item with different parts removed. 64\% of the edited images (289) were produced with the generative editor Nano Banana~2 and 36\% (163) with a manual eraser (cleanup.pictures). All photographs are from Wikimedia Commons, each documented with its URL and licence. \emph{Collection:} photographs were selected by hand so that the removed part is essential to the object's function; every item was verified by three annotators under the rule above, and items that failed the rule were re-edited until they passed. \emph{Preprocessing:} edited images are produced at a resolution that differs from the original for most items (for example a 450$\times$450 original and a 1024$\times$1024 edit); for API calls every image is converted to RGB, limited to 1600~px on its longest side and encoded as JPEG at quality 95. \emph{Uses:} each image and its edit are released under the licence of the source photograph, with edits of share-alike images under the same licence; the intended use is the evaluation of absence detection with the paired original as control. \emph{Distribution:} images, \texttt{metadata.csv} (item id, part, source, licence, editor, source-photograph id), every per-item model output as JSONL, verbatim prompts and scripts. \emph{Maintenance:} the authors maintain the repository; corrections to items or licences are recorded in the metadata file.

\section{Additional results}
\label{app:extra}

\paragraph{Full answer distributions under the detector report.}
Table~\ref{tab:appdist} gives the complete \vis{}/\hs{}/\nv{} distribution on edited images for the pre-answer and post-answer requests, together with the \nv{} rate of the independent first request of the post run, which is the answer that the post request quotes.

\begin{table}[h]
\centering
\small
\begin{tabular}{lrrrrrrr}
\toprule
& \multicolumn{3}{c}{Pre-answer request} & \multicolumn{3}{c}{Post-answer request} & First request \\
\cmidrule(lr){2-4}\cmidrule(lr){5-7}\cmidrule(lr){8-8}
Model & \vis{} & \hs{} & \nv{} & \vis{} & \hs{} & \nv{} & \nv{} \\
\midrule
Claude Sonnet 5.5 & 18.6 & 26.1 & 55.3 & 16.2 & 22.3 & 61.5 & 57.1 \\
Gemini 3.8 Flash & 18.4 & 6.9 & 74.8 & 15.7 & 8.0 & 76.3 & 71.0 \\
GPT-6 Luna & 30.1 & 17.0 & 52.9 & 26.1 & 14.2 & 59.7 & 43.8 \\
GPT-6.1 Sol & 31.2 & 16.4 & 52.4 & 28.3 & 9.1 & 62.6 & 44.9 \\
Grok 4.7 & 26.3 & 13.1 & 60.6 & 27.0 & 9.3 & 63.7 & 47.1 \\
GLM 5.3 FlashX & 20.6 & 22.6 & 56.9 & 20.4 & 17.7 & 61.9 & 46.0 \\
Kimi K2.5 & 37.8 & 6.6 & 55.5 & 20.4 & 4.6 & 75.0 & 45.1 \\
MiMo v2.6 Pro$^{\dagger}$ & 48.9 & 4.0 & 47.1 & 39.6 & 2.7 & 57.7 & 43.6 \\
Qwen3.8 Omni Flash & 17.5 & 19.0 & 63.5 & 11.7 & 11.9 & 76.3 & 58.4 \\
\bottomrule
\end{tabular}
\caption{Answer distributions on the 452 edited images (\%) under the pre-answer and post-answer detector report, and the \nv{} rate of the first request of the post run. Rows sum to 100 up to rounding. $^{\dagger}$First-request rate over the 438 items with a parsed first answer (191 \nv{}); 14 first responses of this model were malformed.}
\label{tab:appdist}
\end{table}

\paragraph{Single-item and multi-item photographs.}
Items from the same photograph share a scene. Table~\ref{tab:appphoto} compares, for each model, the \nv{} rate on items from the 77 photographs that contribute a single item with the rate on items from the 98 photographs that contribute several. The two rates differ by at most 7.8 points, with the sign varying across models.

\begin{table}[h]
\centering
\small
\begin{tabular}{lrrr}
\toprule
Model & Single-item photographs & Multi-item photographs & Difference (pp) \\
\midrule
Claude Sonnet 5.5 & 58.4 & 53.3 & $+5.1$ \\
Gemini 3.8 Flash & 76.6 & 68.8 & $+7.8$ \\
GPT-6 Luna & 42.9 & 42.7 & $+0.2$ \\
GPT-6.1 Sol & 51.9 & 45.9 & $+6.0$ \\
Grok 4.7 & 53.2 & 49.3 & $+3.9$ \\
GLM 5.3 FlashX & 44.2 & 46.7 & $-2.5$ \\
Kimi K2.5 & 50.6 & 44.0 & $+6.6$ \\
MiMo v2.6 Pro & 36.4 & 40.3 & $-3.9$ \\
Qwen3.8 Omni Flash & 66.2 & 59.2 & $+7.0$ \\
\bottomrule
\end{tabular}
\caption{\nv{} rate (\%) on edited images, direct query, for items from photographs that contribute one item ($n=77$) and from photographs that contribute several ($n=375$).}
\label{tab:appphoto}
\end{table}

\paragraph{Run-to-run agreement.}
The direct run and the first request of the post run use the identical prompt and are independent calls. Across the nine models the three-way agreement between the two runs is 68.4--93.6\% and the agreement on \nv{}-versus-not is 77.4--94.5\%; the \nv{} rate moves by $-2.9$ to $+3.9$~pp, and an exact McNemar test \citep{mcnemar_1947} on the paired \nv{} labels gives $p = 0.029$ for Claude Sonnet 5.5 and $p > 0.1$ for every other model (Table~\ref{tab:apprun}). With two runs per model this bounds the variation between two draws; it is not an estimate of the variance over many runs.

\begin{table}[h]
\centering
\small
\resizebox{\linewidth}{!}{\begin{tabular}{lrrrrrr}
\toprule
Model & Direct \nv{} & First request \nv{} & Difference (pp) & \nv{}-vs-not agreement (\%) & Discordant ($+/-$) & McNemar $p$ \\
\midrule
Claude Sonnet 5.5 & 54.2 & 57.1 & $+2.9$ & 93.1 & 22 / 9 & 0.029 \\
Gemini 3.8 Flash & 70.1 & 71.0 & $+0.9$ & 94.2 & 15 / 11 & 0.557 \\
GPT-6 Luna & 42.7 & 43.8 & $+1.1$ & 89.2 & 27 / 22 & 0.568 \\
GPT-6.1 Sol & 46.9 & 44.9 & $-2.0$ & 94.5 & 8 / 17 & 0.108 \\
Grok 4.7 & 50.0 & 47.1 & $-2.9$ & 84.3 & 29 / 42 & 0.154 \\
GLM 5.3 FlashX & 46.2 & 46.0 & $-0.2$ & 77.7 & 50 / 51 & 1.000 \\
Kimi K2.5 & 45.1 & 45.1 & $0.0$ & 84.1 & 36 / 36 & 1.000 \\
MiMo v2.6 Pro$^{\dagger}$ & 39.7 & 43.6 & $+3.9$ & 77.4 & 58 / 41 & 0.107 \\
Qwen3.8 Omni Flash & 60.4 & 58.4 & $-2.0$ & 88.3 & 22 / 31 & 0.272 \\
\bottomrule
\end{tabular}}
\caption{Agreement between two independent runs of the direct query on the 452 edited images. ``Discordant'' counts the items answered \nv{} only in the first request ($+$) or only in the direct run ($-$); $p$ is the exact McNemar test on these counts. Three-way agreement over the nine models lies between 68.4\% and 93.6\%. $^{\dagger}$Over the 438 items on which this model's first request returned a parsed answer (direct \nv{} 174, first-request \nv{} 191); the other eight rows use all 452.}
\label{tab:apprun}
\end{table}

\paragraph{Detector threshold sweep.}
Table~\ref{tab:appsweep} reports, for OWLv2 and YOLOE, the recall of the part on originals and the rejection rate on edited images at a range of confidence thresholds. Lowering the threshold raises recall on originals and lowers the rejection of edited images in step: no threshold separates the two sets.

\begin{table}[h]
\centering
\small
\begin{tabular}{llrr}
\toprule
Detector & Threshold & Original recall (\%) & Edited reject (\%) \\
\midrule
OWLv2-large & 0.05 & 92.7 & 12.8 \\
OWLv2-large & 0.1 & 85.8 & 25.9 \\
OWLv2-large & 0.2 & 64.8 & 53.1 \\
OWLv2-large & 0.3 & 47.8 & 73.7 \\
OWLv2-large & 0.5 & 17.0 & 94.5 \\
\midrule
YOLOE-26l & 0.05 & 53.1 & 59.5 \\
YOLOE-26l & 0.1 & 37.4 & 74.8 \\
YOLOE-26l & 0.2 & 21.5 & 88.7 \\
YOLOE-26l & 0.3 & 12.6 & 92.9 \\
\bottomrule
\end{tabular}
\caption{Threshold sweep on the 452 pairs. Original recall is the share of originals on which the part is detected above the threshold; edited reject is the share of edited images on which it is not.}
\label{tab:appsweep}
\end{table}

\paragraph{Probe AUROC at every layer.}
Table~\ref{tab:applayers} lists the linear-probe AUROC on the residual state of the last prompt token at each of the 37 layers of Qwen3-VL-8B-Instruct (layer 0 is the embedding output, layer 36 the final post-norm state). The permutation null is $0.50 \pm 0.02$.

\begin{table}[h]
\centering
\small
\begin{tabular}{rr@{\hskip 2em}rr@{\hskip 2em}rr@{\hskip 2em}rr}
\toprule
Layer & AUROC & Layer & AUROC & Layer & AUROC & Layer & AUROC \\
\midrule
0 & 0.50 & 10 & 0.92 & 20 & 0.93 & 30 & 0.91 \\
1 & 0.87 & 11 & 0.92 & 21 & 0.94 & 31 & 0.91 \\
2 & 0.90 & 12 & 0.91 & 22 & 0.94 & 32 & 0.90 \\
3 & 0.91 & 13 & 0.91 & 23 & 0.93 & 33 & 0.91 \\
4 & 0.91 & 14 & 0.90 & 24 & 0.93 & 34 & 0.91 \\
5 & 0.92 & 15 & 0.91 & 25 & 0.92 & 35 & 0.89 \\
6 & 0.92 & 16 & 0.91 & 26 & 0.92 & 36 & 0.89 \\
7 & 0.92 & 17 & 0.91 & 27 & 0.92 & & \\
8 & 0.92 & 18 & 0.92 & 28 & 0.92 & & \\
9 & 0.91 & 19 & 0.92 & 29 & 0.91 & & \\
\bottomrule
\end{tabular}
\caption{Linear-probe AUROC (edited versus original) per layer on 448 pairs, PCA-256 and logistic regression, folds grouped by source photograph.}
\label{tab:applayers}
\end{table}

\paragraph{Logit lens at every layer.}
Table~\ref{tab:applens} lists the logit-lens AUROC at the same position: the state of each layer is passed through the final norm and the unembedding rows of the three answer digits, and the renormalised probability of the digit 2 (``not visible'') is used as the score.

\begin{table}[h]
\centering
\small
\begin{tabular}{rr@{\hskip 2em}rr@{\hskip 2em}rr@{\hskip 2em}rr}
\toprule
Layer & AUROC & Layer & AUROC & Layer & AUROC & Layer & AUROC \\
\midrule
0 & 0.50 & 10 & 0.43 & 20 & 0.56 & 30 & 0.29 \\
1 & 0.53 & 11 & 0.51 & 21 & 0.40 & 31 & 0.36 \\
2 & 0.48 & 12 & 0.46 & 22 & 0.38 & 32 & 0.56 \\
3 & 0.53 & 13 & 0.39 & 23 & 0.60 & 33 & 0.65 \\
4 & 0.58 & 14 & 0.41 & 24 & 0.42 & 34 & 0.66 \\
5 & 0.57 & 15 & 0.39 & 25 & 0.64 & 35 & 0.69 \\
6 & 0.55 & 16 & 0.45 & 26 & 0.59 & 36 & 0.75 \\
7 & 0.51 & 17 & 0.46 & 27 & 0.31 & & \\
8 & 0.50 & 18 & 0.47 & 28 & 0.37 & & \\
9 & 0.43 & 19 & 0.51 & 29 & 0.41 & & \\
\bottomrule
\end{tabular}
\caption{Logit-lens AUROC (edited versus original) per layer on the same 448 pairs and position as Table~\ref{tab:applayers}.}
\label{tab:applens}
\end{table}

\paragraph{Steering grid.}
Table~\ref{tab:appsteer} gives the full steering grid on the 179 held-out pairs. With the probe direction added at the output of layer 18, the immediate-answer AUROC remains at 0.825--0.829 at every magnitude while the \nv{} rate on edited images and the false-\nv{} rate on originals rise together. At layers 27 and 33, every magnitude leaves the \nv{} rate within 37.4--39.7\% and the AUROC within 0.827--0.829, a constancy that is expected when the same vector is added to every image. With chain of thought, the class-mean direction at layer 27 raises the \nv{} rate on both edited and original images, and balanced accuracy rises from 62.6\% to 71.7\%; the probe direction at the same layer does not change the reasoned answer. The cosine between the probe direction at the final state and the difference of the unembedding rows of the answer digits 2 and 0 is $-0.03$; two directions in a space of this dimension are close to orthogonal by default, so we draw no conclusion from this value.

\begin{table}[h]
\centering
\small
\resizebox{\linewidth}{!}{\begin{tabular}{lllrrrr}
\toprule
Answer & Direction, layer & $\alpha$ & Edited \nv{} (\%) & Original false \nv{} (\%) & Balanced acc.\ (\%) & AUROC \\
\midrule
Immediate & probe, layer 18 & 0 & 37.4 & 5.6 & 65.9 & 0.828 \\
Immediate & probe, layer 18 & 8 & 40.8 & 6.7 & 67.0 & 0.828 \\
Immediate & probe, layer 18 & 16 & 47.5 & 7.8 & 69.8 & 0.829 \\
Immediate & probe, layer 18 & 32 & 54.7 & 13.4 & 70.7 & 0.825 \\
Immediate & probe, layer 27 & 2--32 & 37.4--38.0 & 5.0--5.6 & 66.2 & 0.827--0.828 \\
Immediate & probe, layer 33 & 2--32 & 38.0--39.7 & 5.0--5.6 & 66.2--67.0 & 0.828 \\
\midrule
Chain of thought & class mean, layer 27 & 0 & 27.9 & 2.8 & 62.6 & -- \\
Chain of thought & class mean, layer 27 & 2 & 36.9 & 5.6 & 65.6 & -- \\
Chain of thought & class mean, layer 27 & 4 & 45.8 & 7.8 & 69.0 & -- \\
Chain of thought & class mean, layer 27 & 8 & 65.2 & 21.8 & 71.7 & -- \\
Chain of thought & probe, layer 27 & 4 & 26.8 & 2.8 & 62.0 & -- \\
Chain of thought & probe, layer 27 & 8 & 26.8 & 2.8 & 62.0 & -- \\
Chain of thought & probe, layer 27 & 16 & 25.1 & 2.8 & 61.2 & -- \\
\bottomrule
\end{tabular}}
\caption{Activation steering on 179 held-out pairs. $\alpha$ is the steering magnitude in units of the class-mean gap along the direction; the largest value for the probe direction corresponds to 27\% of the state norm. AUROC is that of the immediate answer. Chain-of-thought rows report the answer after reasoning, with greedy decoding, as rates over the responses that contain a parsed answer.}
\label{tab:appsteer}
\end{table}

\end{document}